\documentclass[sigconf]{acmart}
\pdfoutput=1

\usepackage{booktabs} 
\usepackage{url}
\usepackage{array,multirow}
\usepackage{textcomp}
\usepackage{graphicx}
\usepackage{subcaption}
\usepackage{tikz}
\newcolumntype{L}{>{\centering\arraybackslash}m{2cm}}
\newcolumntype{N}{>{\centering\arraybackslash}m{5cm}}
\newcolumntype{?}{!{\vrule width 2pt}}

\copyrightyear{2021}
\acmYear{2021}
\setcopyright{none}
\acmConference[SAC '21]{The 36th ACM/SIGAPP Symposium on Applied Computing}{March 22--26, 2021}{Virtual Event, Republic of Korea}

\begin{document}
\title{A Comparison of Similarity Based Instance Selection Methods for Cross Project Defect Prediction}
\renewcommand{\shorttitle}{Comparison of Instance Selection Methods for CPDP}

\author{Seyedrebvar Hosseini}
\authornote{S .Hosseini is also with F-secure, Helsinki, Finland.}
\affiliation{%
  \institution{M3S Research Unit, Faculty of ITEE\\ University of Oulu}
  \city{Oulu, Finland}
}
\email{rebvar@oulu.fi}

\author{Burak Turhan}
\authornote{B. Turhan is also an Adjunct Professor (Research) at the Faculty of IT, Monash University, Australia.}
\affiliation{%
  \institution{ M3S Research Unit, Faculty of ITEE \\University of Oulu}
  \city{Oulu, Finland}
}
\email{burak.turhan@oulu.fi}

\begin{abstract}
    \textbf{Context}: Previous studies have shown that training data instance selection based on nearest neighborhood (NN) information can lead to better performance in cross project defect prediction (CPDP) by reducing heterogeneity in training datasets. However, neighborhood calculation is computationally expensive and approximate methods such as Locality Sensitive Hashing (LSH) can be as effective as exact methods.
    \textbf{Aim}: We aim at comparing instance selection methods for CPDP, namely LSH, NN-filter, and Genetic Instance Selection (GIS). 
    \textbf{Method}: We conduct experiments with five base learners, optimizing their hyper parameters, on 13 datasets from PROMISE repository in order to compare the performance of LSH with benchmark instance selection methods NN-Filter and GIS.
    \textbf{Results}: The statistical tests show six distinct groups for F-measure performance. The top two group contains only LSH and GIS benchmarks whereas the bottom two groups contain only NN-Filter variants. LSH and GIS favor recall more than precision. In fact, for precision performance only three significantly distinct groups are detected by the tests where the top group is comprised of NN-Filter variants only. Recall wise, 16 different groups are identified where the top three groups contain only LSH methods,  four of the next six are GIS only and the bottom five contain only NN-Filter. Finally, NN-Filter benchmarks never outperform the LSH counterparts with the same base learner, tuned or non-tuned. Further, they never even belong to the same rank group, meaning that LSH is always significantly better than NN-Filter with the same learner and settings. 
    \textbf{Conclusions}: The increase in performance and the decrease in computational overhead and runtime make LSH a promising approach. However, the performance of LSH is based on high recall and in environments where precision is considered more important NN-Filter should be considered.
\end{abstract}

%



\maketitle

\section{Introduction}

The key premise of cross project defect prediction (CPDP) is learning from projects that are different than those to which the models are applied \cite{A9,A21}. Hence, in presence of {\em relevant} data from other projects, including open source ecosystems, CPDP can result in practical applications for making predictions for new projects.
 
 The number of studies addressing the challenges of CPDP through different strategies has been increasing in the last decade \cite{ourslr}. However, one big challenge of CPDP, which is how to identify the aforementioned {\em relevant} data instances, still remains an open problem. In that respect, a widely used strategy is to use row processing methods with the goal of sampling a {\em relevant} subset of training data instances from the available corpus \cite{ourslr}. Among row processing methods, relevancy filtering \cite{A21} has become a standard benchmark in CPDP studies and is {\em ``nearly always among the best"} \cite{Herbold17}. 
 
 In this paper, we investigate an alternative relevant data selection method based on hashing, namely locality sensitive hashing (LSH). LSH has two advantages over conventional relevancy filtering. First, LSH is fast, like most hash based operations, and it provides the option of finding similar items in a large set without the cost of examining each pair, unlike the conventional nearest neighbor search. Second, the hashing method is an approximate calculation, introducing a degree of randomness in its approach to leverage exploration (over exploitation) for selecting neighboring data points. This is a desired aspect in recently proposed search based methods for relevant data selection for CPDP, i.e., GIS \cite{gis1,gis2}, considering potential data quality issues in defect data. Accordingly, we use GIS as another benchmark method from the search-based family, fully utilizing the exploration vs. exploitation aspects. 
 
 Accordingly, the aim of this study is to answer the following research question: \\

\textbf{RQ}: How is the performance of LSH in comparison with the benchmark methods, i.e., relevancy filtering and search-based data selection?\\

We conduct a set of experiments with five base learners (Naive Bayes, Logistic Regression, Decision Table, Decision Tree and Bayes Net), optimizing their hyper parameters (using a modified variation of grid search), on 13 datasets from PROMISE repository in order to compare the performance of the proposed method, LSH, with benchmark instance selection methods, namely NN-Filter and GIS. We find that LSH is always significantly, though marginally, better than NN-Filter with the same learner and settings yet with the additional advantages of reduced runtime and better scalability.

The scope of our study is limited to the comparison of neighborhood selection methods using standard machine learning algorithms as base learners. Hence, improving the state of the art CPDP learners \cite{ourslr} and a comparison with within project defect prediction (WPDP) is out of our scope. 


This paper is organized as follows: The next section summarizes the related studies on CPDP and briefly describes how our study differs. Proposed approach, datasets and experimental procedures are presented in Section \ref{sec:mtd}. Section \ref{sec:res} presents the results of our analysis and discussions. Section \ref{sec:thr} discusses the threats to the validity of our study. Finally, the last section concludes the paper with a summary of the findings and directions for future work.

\section{Related Work}
\label{sec:rltd}
The systematic literature review by Hosseini et al \cite{ourslr} presents detailed discussions of the state of the art CPDP approaches in the literature with a specific focus on data approaches used. We describe the most relevant studies to our work next.

Turhan et al. \cite{A21} found that despite its good probability of detection rates CPDP causes excessive false alarms. To overcome this problem, they proposed NN-Filter to select the most relevant training data instances from a pool of cross project datasets. Even though this method lowered the false alarm rates dramatically, its performance was still worse than WPDP.

Zimmermann et al. \cite{A9} tested the CPDP approach for 622 pairs of 28 datasets from 12 projects (open source and commercial) and found only 21 pairs (3.4\%) that match their performance criteria (precision, recall and accuracy, all greater than 0.75). This means that the predictions will fail in most cases if training data is not selected carefully. They also found that CPDP is not symmetrical as data from Firefox can predict Internet Explorer defects but the opposite does not hold. They argued that characteristics of data and process are crucial factors for CPDP.

He et al. \cite{A5} proposed to use the distributional characteristics (median, mean, variance, standard deviation, skewness, quantiles, etc.) for training dataset selection. They conclude that in the best cases cross project data may provide acceptable prediction results. They also state that training data from the same project does not always lead to better predictions and carefully selected cross project data may provide better prediction results than within-project (WP) data. They also found that data distributional characteristics are informative for training data selection.

Herbold \cite{A24} proposed distance-based strategies for the selection of training data based on distributional characteristics of the available data. They presented two strategies based on EM (Expectation Maximization) clustering and NN (Nearest Neighbor) algorithm with distributional characteristics as the decision strategy. They observed that i) weights can be used to successfully deal with biased data and ii) the training data selection provides a significant improvement in the success rate and recall of defect detection. However, their overall success rate was still too low for the practical application of CPDP.

Turhan et al. \cite{A13} evaluated the effects of mixed project data on predictions. They tested whether mixed WP and CP data improves the prediction performances. They performed their experiments on 73 versions of 41 projects using Na\"{\i}ve Bayes classifier. They concluded that the mixed project data would significantly improve the performance of the defect predictors.

Ryu et al. \cite{A45} presented a Hybrid Instance Selection with the Nearest Neighbor (HISNN) method using a hybrid classification to address the class imbalance for CPDP. Their approach used a combination of the Nearest Neighbour algorithm with Hamming distance and Na\"{\i}ve Bayes to address the instance selection problem.

Instance selection and data quality are investigated by Hosseini et al. \cite{gis1,gis2}. The proposed model, called GIS, approaches the instance selection problem using a search based method based on genetic algorithm and tries to build evolving training datasets while utilizing NN-Filter in the dataset evaluation process. The model also takes steps toward addressing the data quality issues in the defect datasets.  Hosseini et al. further investigated design parameter selection in the context of GIS models. Beside the inferiority of the relevancy filtering of NN-Filter in performance to GIS, they asserted that the performance increase due to incorporation of NN-Filter is not significant. Further, being arguably slow itself, NN-Filter makes GIS, which utilized NN-Filter, more time consuming with increases in the size of the pool of available data. We address these issues by utilizing Locality Sensitive Hashing (LSH) as an alternative. Next section provides the details of LSH, GIS and NN-Filter.

Our work is different from the above-mentioned studies in two ways. First, studies in CPDP rarely have considered approaches that target multiple aspects of the models, e.g., both of instance selection and parameter optimization. Second, practicality of building and applying the models are usually neglected which for some models (e.g., meta-learner in \cite{A5}), especially instance selection methods in large dataset contexts are necessary. Please notice that comparison with within project benchmarks is not in the scope of this study, because we focus on instance selection methods in large dataset contexts which is not applicable in the case of within project data. 

\section{Research Method}
\label{sec:mtd}

This section describes the details of our study. We present the proposed approach as well as the benchmark methods, the datasets and metrics, and the performance evaluation criteria used in our study.

\subsection{Motivation}

In our earlier studies, we presented empirical evidence for the usefulness of careful data selection by designing a genetic instance selection approach, called GIS, for which we observed superior performances to those of CPDP benchmarks and comparable and superior to those from WPDP benchmarks. Of the reasons for the observed performance increase, incorporating NN-Filter as validation dataset selection method can be pointed out. Another advantage of GIS is that it tries to address the quality issues \cite{Shepperd13} which are present in defect prediction datasets.

The CPDP benchmark NN-Filter, which was used as validation dataset selection method for GIS, selects instances based on a defined distance function (e.g., euclidean distance). The performance increases for GIS variants over NN-Filter were motivations to investigate alternatives to NN-Filter. The fundamental difference of CPDP and WPDP is another reason why we seek other/better alternatives to NN-Filter. In CPDP domain, the problem of lack of available data in WPDP shifts to the lack of relevant data due to the abundance of data from other projects. Due to available CPDP datasets, scarcity of training data diminishes and instead scarcity of useful training data arises due to the dataset shift problem, for which CPDP is a clear example \cite{A56}. 

In presence of many available CPDP dataset, the relevancy filtering of NN-Filter not only is inferior to GIS but also becomes very slow itself and makes GIS more time consuming as the size of the pool of available cross project data increases. This is where the alternative near neighbor selection method, i.e., Locality Sensitive Hashing (LSH) can be considered as an alternative.

\subsection{Proposed Approach}

Locality Sensitive Hashing (LSH) is a family of hashing methods that produce the same hash for similar vectors \cite{lsh0}. LSH is a member of approximate near neighbor search methods that was primarily proposed to tackle the curse of dimensionality of exact NN searches \cite{lsh0}. The idea in LSH is that rather than retrieving the exact closest neighbor,
one can make a ``good guess'' of the nearest neighbor. LSH aims at mapping close points to the same buckets while putting faraway points into different buckets. Therefore, one needs to only examine the items that are mapped to the same bucket. Hence, if designed correctly, only a small fraction of pairs need examination. On the other hand, the process might generate false negatives. 

LSH has been applied in a wide range of areas including near-duplicate detection \cite{lsh7}, hierarchical clustering \cite{lsh9}, image similarity identification \cite{lsh13}, audio similarity identification and fingerprint \cite{lsh14audio}, digital video fingerprinting, spam detection and nearest neighbor search \cite{booklsh}.

Different LSH functions exist for various similarity metrics. Example hash functions include MinHash algorithm designed for Jaccard similarity (the relative number of elements that two sets have in common), random projection, super-bit for cosine similarity and bit sampling for Hamming distance \cite{booklsh}. 

There are two main use cases for LSH functions, (1) computing the signature of large input vectors which can be used for quick similarity estimation of input vectors and  (2) binning similar vectors together with a given number of buckets.

The LSH algorithm based on random projection \cite{lsh2,lsh132} is developed to solve the near neighbor search problem under the angle between two vectors. Super-bit is an improvement on Random Projection LSH and relies on an estimation of cosine similarity. In super-bit, the $K$ random vectors are orthogonalized in $L$ batches of $N$ vectors, where $N$ is called the super-bit depth, $L$ is called the number of super-bits, $K = L \times N$ is the code length, or the size of the signature \cite{lshsuperbit}.

The cosine similarity between two vectors in $R^n$ is the cosine of their angle and is computed as 
\begin{equation}
v1 . v2 / (|v1| \times |v2|)
\end{equation}

Cosine similarity is a measure of similarity between two non-zero vectors of an inner product space that measures the cosine of the angle between them. The cosine of angle 0 is 1, and it is less than 1 for any other angle in the interval $[0,2\pi)$. It is, therefore, a judgment of orientation and not magnitude.

We use LSH to split the training data into smaller subsets (buckets) which are expected to be in close proximate of each other. Such subsets in turn can act as training data themselves. We select the best bucket by evaluating each bucket on a set of randomly selected validation datasets. The average fitness assigned to each bucket is used to determine the best bucket and  it will be used as the final training data. The division process is  similar to clustering the data into distinct groups and selecting the best cluster (by assigning a fitness value), and using it as training data. One advantage of using LSH is that the method is fast and suitable for mining massive datasets both instance and feature wise \cite{lsh0, booklsh}.

\subsection{Benchmarks}
We use GIS and NN-Filter as benchmarks for the proposed LSH based data selection method. Descriptions of these approaches are presented next. 

\subsubsection{GIS}

This section presents a brief description of the search based method which is used in this study, namely Genetic Instance Selection (GIS) proposed in previous studies \cite{gis1,gis2}. To make predictions for a test set, GIS starts by generating random subsets of instances belonging to the pool of instances from other project. A validation dataset is generated, initial training datasets from the first population are evaluated on the validation dataset, and fitness values are calculated. Using the genetic operations, e.g., mutation and crossover, new generations are created and best performing datasets survive and are transferred to the next generations. The original GIS approach used fixed length training datasets, NN-Filter as validation dataset generation method and Naive Bayes (NB) as learner. The mutation and cross over operations are designed to address the data quality and dataset shift problem to some extent through random label changes in training data, accounting for the potential noise in the data. GIS also incorporates an alternative validation data method. Beside the NN-Filter generated data, another version of GIS is designed to use multiple randomly generated datasets for validation and evaluations. Average fitness of each dataset acts as the basis for generating new populations in such settings. The population size is 40 and the maximum number of generations is 20 by default.  The reason for selecting these values is due to having small population sizes which cover almost all of the original training instances in most iterations (95\%) of the initial populations despite the small sizes. Further, the process converges quickly, justifying the selection of the maximum number of generations. Both Fixed (FX) and Variable(VR) training dataset size generation methods are considered. For FX, one percent of the pool of instances are selected initially and are kept the at the same size by the genetic operators. For VR, the initial datasets can range from minimum of 100 instances to maximum of the entire pool of training data from other projects. Cross over operation uses two point cross over to account for variable chromosome sizes.

\subsubsection{NN-Filter}: In this approach, the most relevant training instances are selected based on a distance measure \cite{A21}, e.g., the k-nearest neighbours using Euclidean distance. This approach has been used and validated in CPDP \cite{A45,A46,A21}. Further, GIS uses NN-Filter to select the validation dataset. NN-Filter with different values for k has been used in the literature \cite{ourslr}. Instead of selecting a specific value for k, however, we try to find the optimal k value from the range of 1 to 10, using an evaluation process. Finally, we use a constant 10-NN-Filter in GIS as in the original NN-filter study settings with nearest neighbor as validation data.

\subsection{Datasets and Metrics}

We used 13 projects from the PROMISE repository for our experiments. 12 of these projects are open source and one of them is a proprietary project (prop-6). These datasets are collected by Jureczko, Madeyski and Spinellis \cite{A23,A50}. The list of the datasets is presented in Table \ref{tbldatasets} with the corresponding size and defect information. The reason for using these datasets is driven by our goal to account for noise in the data which is a threat specified by the donors of these datasets. Each dataset contains a number of instances corresponding to the classes in the release. Each instance has 20 static code metrics. The list of these metrics are presented in Table \ref{tblmetrics}. We use the full set of metrics for all benchmarks.

\begin{table}[t]
\setlength\tabcolsep{0.10cm}
\fontsize{9}{9}\selectfont
  \centering
  \caption{Datasets used in the study. DP= Defect Prone. }
    \begin{tabular}{lllllllllll} \hline\hline
    \textbf{Release} & Abbr. & \#CLS & \#DP & DP(\%) & \#LOC \\ \hline\hline
    \textbf{ant-1.7} & ANT & 745   & 166   & 22.3 & 209K \\ \hline 
    \textbf{camel-1.6} & CML & 965   & 188   & 19.5 & 113K  \\ \hline
    \textbf{ivy-2.0} & IVY & 352   & 40    & 11.4 & 88K  \\ \hline
    \textbf{jedit-4.3}& jED & 492   & 11    & 02.2 & 202K  \\ \hline
    \textbf{log4j-1.2}& L4J & 205   & 189   & 92.2 & 38K \\ \hline
    \textbf{lucene-2.4}& LUC & 340   & 203   & 59.7 & 103K \\ \hline
    \textbf{poi-3.0}& POI & 442   & 281   & 63.6 & 129K  \\ \hline
    \textbf{prop-6.0}& PR6 & 660   & 66   & 10.0 & 98K  \\ \hline
    \textbf{synapse-1.2}& SYN & 256   & 86   & 33.6 & 54K  \\ \hline
    \textbf{tomcat-6.0}& TOM & 885   & 77   & 09.0 & 301K  \\ \hline
    \textbf{velocity-1.6}& VEL & 229   & 78   & 34.1 & 57K  \\ \hline
    \textbf{xalan-2.7}& XAL & 885   & 411   & 46.4 & 412K  \\ \hline
    \textbf{xerces-1.4}& XER & 588   & 437   & 74.3 & 141K  \\ \hline
    \end{tabular}%
  \label{tbldatasets}%
\end{table}

\begin{table}[t]
\setlength\tabcolsep{0.10cm}
\fontsize{9}{9}\selectfont
  \centering
  \caption{List of the metrics used in this study}
    \begin{tabular}{lp{1.3cm}p{5.5cm}} \hline
    \textbf{ID} &\textbf{Variable} & \textbf{Description} \\\hline\hline
    1 & WMC   & Weighted Methods per Class \\\hline
    2 & DIT   & Depth of Inheritance Tree \\\hline
    3 & NOC   & Number of Children \\\hline
    4 & CBO   & Coupling between Object classes \\\hline
    5 & RFC   & Response for a Class \\\hline
    6 & LCOM  & Lack of Cohesion in Methods \\\hline
    7 & CA  &  Afferent Couplings \\\hline
    8 & CE & Efferent Couplings\\\hline    
    9 & NPM & Number of Public Methods\\\hline
    10 & LCOM3 & Normalized version of LCOM\\\hline
    11 & LOC   & Lines Of Code \\\hline
    12 & DAM & Data Access Metric\\\hline
    13 & MOA & Measure Of Aggregation\\\hline
    14 & MFA & Measure of Functional Abstraction\\\hline
    15 & CAM & Cohesion Among Methods\\\hline
    16 & IC & Inheritance Coupling\\\hline
    17 & CBM & Coupling Between Methods\\\hline
    18 & AMC & Average Method Complexity\\\hline
    19 & MAX\_CC  & Maximum cyclomatic complexity\\\hline
    20 & AVG\_CC &  Mean cyclomatic complexity\\\hline
    \end{tabular}%
  \label{tblmetrics}%
\end{table}%

\subsection{Performance Measures and Tools}

We use Precision, Recall, F-measure and GMean for evaluations, optimization, and performance reporting. These indicators are calculated by comparing the outcome of the prediction models and the actual label of the data instances. Using the generated confusion matrices, one can extract a diverse set of performance indicators, the mentioned set of which are used in this study. F-Measure is selected in this study as the basis of our selection of the best approach. Additionally, a combination of F-measure and GMean, i.e., F$\times$GMean is used further as the basis for fitness assignment for LSH parameter tuning, learning technique hyper-parameter tuning, and finally as the fitness function for GIS. The same fitness function is used in previous studies \cite{gis1,gis2, gis3}.

All the experiments are conducted using \textbf{WEKA}\footnote{\url{http://www.cs.waikato.ac.nz/ml/weka/}} machine Learning tool version 3.8.1. The relevant WEKA classes for instance manipulation are modified for the data related operations. The statistical tests are carried out using the \textbf{scipy.stats}\footnote{\url{https://www.scipy.org/}} library for python, and \textbf{Scott-Knott-ESD \cite{sctnt}}\footnote{\url{https://rdrr.io/cran/ScottKnottESD/}} library for \textbf{R}. The plots are generated using \textbf{R} as well. For LSH, we use the \textbf{JavaLSH} implementation, which is open source and is available online\footnote{\url{https://github.com/tdebatty/java-LSH/}}. The code and scripts for the experiments as well as the modified WEKA version are available online to ease the replication and for further validation \cite{replpack}. 

To measure the performance of the methods and comparisons, the Scott-Knott Effect Size Difference (ESD) tests are used.  The Scott-Knott ESD test is a mean comparison approach based on hierarchical clustering to partition the set of treatment means into statistically distinct groups with non-negligible difference \cite{sctnt}. This version of Scott-Knott test considers the magnitude of the differences (effect sizes) of treatment means within a group and between groups. The Scott-Knott ESD test produces the ranking of treatment means while ensuring that (1) the magnitude of the differences for all of the treatments in the same group are negligible and (2) the magnitude of the differences of treatments between groups are non-negligible \cite{sctnt}. We use Cliff's delta \cite{clf} effect size to interpret the magnitude of differences and to compare the identified groups instead of individual benchmarks/methods.

\subsection{Learning Techniques}

We use five learning techniques in this study for all benchmarks. These learners are Na\"{\i}ve Bayes (NB), BayesNet (BN), Logistic Regression (LOG), DecisionTable (DT) and DecisionTree (J48). These learners belong to four different families of learning techniques and their performances in general and in particular for defect prediction have been demonstrated in previous studies \cite{ourslr,A55,A71,Less1}. Being very fast and having randomness in its settings, LSH method provides the opportunity of tuning the hyper-parameters of the learners in a reasonable fashion.  We performed hyper-parameter tuning for each learning technique (NB\{K:Use kernel density estimator, D: Use supervised discretization\}, DT\{E:Performance evaluation measure, I: Use nearest neighbour instead of global table majority, S: Size of lookup cache for evaluated subsets , X: Use cross validation to evaluate features\}, J48\{M: minimum number of instances per leaf, C: confidence threshold for pruning\}, BN\{Q: Search algorithm, D: use of ADTree data structure\}, LOG\{R: ridge in the log-likelihood\}). Default learner settings are used if certain used combinations are not valid. Details of these parameters are available online in the official WEKA machine learning library documentations. 

\subsection{Tuning and Validation Datasets}

To find the best set of parameters, multiple validation datasets are generated randomly in each iteration which are used for performance evaluation. In this variation of ``grid search'' method, the average performance (average F$\times$GMean) is assigned as fitness to each set of candidate parameters and the best settings for the learners are determined. Similarly, these fitness functions are used for LSH parameter tuning, which in some benchmarks involve learner optimization, for each set of parameters. In case of NN-Filter, the assigned fitness values determine the optimal value for k. We utilized 20 randomly generated dataset in each iteration. The datasets are fed into all methods that use them for optimization (variants of LSH and NN-Filter). The size of these datasets are selected randomly in the range of $[125, 250]$. The datasets are guaranteed to include instances from both classes, i.e., including defective and non-defective instances, similar to the defect datasets. 

\subsection{Time Measurement}

One of the goals for using alternative near neighbor selection approaches, and especially approximate methods, is practical considerations. The larger the dataset gets the more impractical the use of exact methods become. With that in mind, we collect elapsed time for each instance selection approach separately. Please notice that we are interested in the instance selection part of each approach and not the actual final predictions. Hence, for NN-Filter, the elapsed time is the time required to select the nearest k instances to each test set instance. Finding optimal k value falls into the same category and requires calculation of NN-Filter for k=1, k=2,..., k=n. Similarly, For LSH the runtime is the amount of time to create the buckets. Having randomization and tuning for optimal parameters for LSH, the buckets need to be created multiple times in each iteration and the elapsed time for them are considered in the reported runtimes. The story is somewhat different for GIS, since GIS needs to evaluate the population members to generate the next population. The process of finding optimal datasets in GIS has a mandatory step, i.e., fitness assignment. Therefore, the elapsed time in GIS is comprised of random initial population generation, genetic operators (selection, cross over and mutation), combining and sorting the populations, and fitness assignment for each population. In case of GIS with NN-Filter as validation dataset, the time for generating the validation dataset is also included in the runtime. We do not report the runtimes of tuning the learning technique hyper-parameters in any of the methods. In fact, GIS does not include learning technique optimization in its settings at all. The experiments are conducted on a shared server. For practical purposes, the experiments for all datasets are performed in parallel. Hence, 13 threads in the thread pool executed the experiments simultaneously for 13 datasets. 

\begin{table*}[!t]
\setlength\tabcolsep{0.10cm}
\fontsize{9}{9}\selectfont
  \centering
  \caption{Median (sorted) F-measure performance per dataset per benchmark. The last column shows runtime in seconds}
    \begin{tabular}{>{\bfseries}llllllllllllll?l|l|l}
    \hline
Method / Dataset & ANT & CML & IVY & jED & L4J & LUC & POI & PR6 & SYN & TOM & VEL & XAL & XER & Med. & Avg$\pm$Std & Time(sec.) \\\hline\hline

NNF-TunedLOG &  0.30  &  0.21  &  0.34  &  0.06  &  0.26  &  0.43  &  0.29  &  0.18  &  0.47  &  0.24  &  0.34  &  0.44  &  0.30  & 0.30 & 0.30$\pm$0.11 & 6.8 \\\hline 
NNF-LOG &  0.33  &  0.21  &  0.35  &  0.06  &  0.27  &  0.46  &  0.29  &  0.18  &  0.34  &  0.22  &  0.34  &  0.45  &  0.32  & 0.32 & 0.29$\pm$0.10 & 6.8 \\\hline 
NNF-NB &  0.48  &  0.19  &  0.37  &  0.06  &  0.13  &  0.33  &  0.28  &  0.23  &  0.53  &  0.34  &  0.33  &  0.46  &  0.27  & 0.33 & 0.31$\pm$0.13 & 6.4 \\\hline 
NNF-J48 &  0.36  &  0.28  &  0.32  &  0.04  &  0.43  &  0.53  &  0.70  &  0.23  &  0.47  &  0.23  &  0.37  &  0.53  &  0.56  & 0.37 & 0.39$\pm$0.17 & 5.8 \\\hline 
NNF-TunedDT &  0.40  &  0.32  &  0.34  &  0.04  &  0.37  &  0.40  &  0.69  &  0.24  &  0.58  &  0.18  &  0.31  &  0.57  &  0.53  & 0.37 & 0.38$\pm$0.17 & 6.9 \\\hline 
NNF-TunedJ48 &  0.33  &  0.27  &  0.27  &  0.03  &  0.51  &  0.53  &  0.48  &  0.23  &  0.48  &  0.22  &  0.38  &  0.52  &  0.58  & 0.38 & 0.37$\pm$0.16 & 5.8 \\\hline 
NNF-DT &  0.40  &  0.27  &  0.38  &  0.05  &  0.26  &  0.38  &  0.78  &  0.26  &  0.58  &  0.31  &  0.30  &  0.43  &  0.53  & 0.38 & 0.38$\pm$0.17 & 6.9 \\\hline 
LSH-J48 &  0.36  &  0.29  &  0.26  &  0.04  &  0.52  &  0.58  &  0.72  &  0.20  &  0.42  &  0.19  &  0.43  &  0.51  &  0.53  & 0.39 & 0.41$\pm$0.22 & 1.0 \\\hline 
LSH-TunedJ48 &  0.36  &  0.29  &  0.23  &  0.05  &  0.57  &  0.56  &  0.71  &  0.23  &  0.44  &  0.16  &  0.42  &  0.55  &  0.57  & 0.40 & 0.41$\pm$0.22 & 1.0 \\\hline 
GIS(VR-VNN)-J48 &  0.38  &  0.30  &  0.26  &  0.05  &  0.53  &  0.55  &  0.70  &  0.21  &  0.45  &  0.22  &  0.42  &  0.61  &  0.60  & 0.41 & 0.40$\pm$0.19 & 20.1 \\\hline 
GIS(VR-VMUL)-J48 &  0.37  &  0.27  &  0.25  &  0.04  &  0.51  &  0.74  &  0.61  &  0.21  &  0.46  &  0.21  &  0.46  &  0.65  &  0.59  & 0.41 & 0.41$\pm$0.20 & 21.0 \\\hline 
LSH-TunedDT &  0.36  &  0.34  &  0.20  &  0.04  &  0.96  &  0.75  &  0.78  &  0.19  &  0.44  &  0.16  &  0.51  &  0.99  &  0.85  & 0.43 & 0.49$\pm$0.30 & 1.0 \\\hline 
NNF-TunedNB &  0.50  &  0.32  &  0.39  &  0.05  &  0.36  &  0.63  &  0.53  &  0.20  &  0.60  &  0.23  &  0.44  &  0.55  &  0.46  & 0.44 & 0.41$\pm$0.16 & 6.4 \\\hline 
GIS(FX-VNN)-NB &  0.42  &  0.33  &  0.32  &  0.04  &  0.69  &  0.70  &  0.81  &  0.20  &  0.55  &  0.20  &  0.37  &  0.78  &  0.66  & 0.44 & 0.47$\pm$0.24 & 30.1 \\\hline 
GIS(FX-VMUL)-NB &  0.44  &  0.29  &  0.36  &  0.06  &  0.44  &  0.72  &  0.76  &  0.20  &  0.50  &  0.30  &  0.43  &  0.91  &  0.77  & 0.44 & 0.47$\pm$0.24 & 54.2 \\\hline 
GIS(VR-VMUL)-NB &  0.44  &  0.27  &  0.35  &  0.06  &  0.45  &  0.73  &  0.77  &  0.20  &  0.50  &  0.31  &  0.44  &  0.86  &  0.81  & 0.44 & 0.48$\pm$0.24 & 54.1 \\\hline 
NNF-TunedBN &  0.44  &  0.26  &  0.33  &  0.07  &  0.50  &  0.48  &  0.63  &  0.29  &  0.48  &  0.22  &  0.42  &  0.58  &  0.59  & 0.44 & 0.41$\pm$0.16 & 6.3 \\\hline 
LSH-DT &  0.36  &  0.34  &  0.20  &  0.04  &  0.96  &  0.75  &  0.79  &  0.19  &  0.48  &  0.16  &  0.51  &  0.99  &  0.85  & 0.45 & 0.50$\pm$0.30 & 1.0 \\\hline 
GIS(FX-VMUL)-J48 &  0.40  &  0.32  &  0.26  &  0.04  &  0.69  &  0.75  &  0.77  &  0.22  &  0.53  &  0.19  &  0.46  &  0.80  &  0.74  & 0.46 & 0.47$\pm$0.24 & 5.0 \\\hline 
GIS(VR-VMUL)-LOG &  0.44  &  0.30  &  0.33  &  0.05  &  0.70  &  0.72  &  0.74  &  0.22  &  0.51  &  0.22  &  0.43  &  0.79  &  0.71  & 0.46 & 0.47$\pm$0.23 & 18.3 \\\hline 
GIS(VR-VNN)-DT &  0.38  &  0.34  &  0.20  &  0.05  &  0.54  &  0.70  &  0.79  &  0.19  &  0.53  &  0.16  &  0.53  &  0.80  &  0.67  & 0.46 & 0.45$\pm$0.25 & 19.1 \\\hline 
LSH-TunedBN &  0.36  &  0.34  &  0.20  &  0.04  &  0.96  &  0.47  &  0.78  &  0.19  &  0.50  &  0.18  &  0.51  &  0.99  &  0.64  & 0.46 & 0.48$\pm$0.29 & 1.1 \\\hline 
GIS(FX-VMUL)-LOG &  0.42  &  0.30  &  0.26  &  0.04  &  0.77  &  0.72  &  0.74  &  0.22  &  0.52  &  0.21  &  0.47  &  0.83  &  0.74  & 0.47 & 0.48$\pm$0.25 & 11.8 \\\hline 
GIS(VR-VNN)-LOG &  0.43  &  0.31  &  0.30  &  0.04  &  0.58  &  0.66  &  0.75  &  0.23  &  0.54  &  0.21  &  0.49  &  0.73  &  0.66  & 0.47 & 0.45$\pm$0.22 & 13.9 \\\hline 
NNF-BN &  0.51  &  0.32  &  0.33  &  0.06  &  0.43  &  0.63  &  0.63  &  0.27  &  0.48  &  0.31  &  0.53  &  0.62  &  0.58  & 0.48 & 0.44$\pm$0.17 & 6.4 \\\hline 
GIS(FX-VNN)-J48 &  0.38  &  0.34  &  0.25  &  0.05  &  0.64  &  0.68  &  0.78  &  0.19  &  0.52  &  0.22  &  0.52  &  0.80  &  0.70  & 0.48 & 0.46$\pm$0.24 & 3.4 \\\hline 
GIS(VR-VNN)-NB &  0.40  &  0.34  &  0.26  &  0.04  &  0.78  &  0.73  &  0.79  &  0.19  &  0.54  &  0.18  &  0.51  &  0.79  &  0.68  & 0.50 & 0.48$\pm$0.26 & 30.6 \\\hline 
GIS(FX-VNN)-LOG &  0.42  &  0.31  &  0.24  &  0.04  &  0.73  &  0.69  &  0.77  &  0.22  &  0.54  &  0.20  &  0.52  &  0.81  &  0.70  & 0.50 & 0.47$\pm$0.25 & 8.6 \\\hline 
GIS(FX-VMUL)-BN &  0.38  &  0.34  &  0.20  &  0.05  &  0.96  &  0.56  &  0.79  &  0.19  &  0.50  &  0.30  &  0.52  &  0.77  &  0.63  & 0.50 & 0.48$\pm$0.26 & 12.5 \\\hline 
GIS(VR-VMUL)-BN &  0.40  &  0.34  &  0.20  &  0.06  &  0.96  &  0.54  &  0.79  &  0.26  &  0.51  &  0.32  &  0.53  &  0.71  &  0.53  & 0.50 & 0.48$\pm$0.24 & 15.9 \\\hline 
LSH-BN &  0.36  &  0.34  &  0.20  &  0.04  &  0.96  &  0.72  &  0.78  &  0.19  &  0.50  &  0.16  &  0.51  &  0.94  &  0.83  & 0.50 & 0.50$\pm$0.30 & 1.1 \\\hline 
LSH-LOG &  0.38  &  0.33  &  0.20  &  0.04  &  0.96  &  0.75  &  0.78  &  0.18  &  0.52  &  0.16  &  0.52  &  0.95  &  0.83  & 0.50 & 0.51$\pm$0.30 & 1.0 \\\hline 
LSH-TunedNB &  0.36  &  0.33  &  0.20  &  0.05  &  0.96  &  0.75  &  0.78  &  0.18  &  0.50  &  0.16  &  0.51  &  0.90  &  0.84  & 0.50 & 0.50$\pm$0.30 & 1.0 \\\hline 
LSH-NB &  0.37  &  0.34  &  0.22  &  0.04  &  0.91  &  0.75  &  0.79  &  0.18  &  0.51  &  0.17  &  0.52  &  0.93  &  0.80  & 0.50 & 0.50$\pm$0.29 & 1.1 \\\hline 
LSH-TunedLOG &  0.36  &  0.33  &  0.20  &  0.04  &  0.95  &  0.75  &  0.78  &  0.18  &  0.51  &  0.17  &  0.52  &  0.94  &  0.84  & 0.50 & 0.50$\pm$0.30 & 1.0 \\\hline 
GIS(VR-VMUL)-DT &  0.44  &  0.34  &  0.20  &  0.05  &  0.96  &  0.65  &  0.79  &  0.25  &  0.59  &  0.28  &  0.53  &  0.71  &  0.59  & 0.52 & 0.48$\pm$0.24 & 29.5 \\\hline 
GIS(FX-VMUL)-DT &  0.38  &  0.34  &  0.20  &  0.05  &  0.90  &  0.72  &  0.77  &  0.22  &  0.53  &  0.26  &  0.53  &  0.78  &  0.65  & 0.52 & 0.48$\pm$0.25 & 9.5 \\\hline 
GIS(FX-VNN)-DT &  0.38  &  0.34  &  0.20  &  0.05  &  0.89  &  0.72  &  0.80  &  0.19  &  0.52  &  0.16  &  0.53  &  0.83  &  0.70  & 0.52 & 0.47$\pm$0.28 & 7.7 \\\hline 
GIS(FX-VNN)-BN &  0.38  &  0.34  &  0.20  &  0.05  &  0.89  &  0.70  &  0.79  &  0.19  &  0.52  &  0.19  &  0.53  &  0.81  &  0.69  & 0.52 & 0.48$\pm$0.27 & 7.0 \\\hline 
GIS(VR-VNN)-BN &  0.38  &  0.34  &  0.21  &  0.05  &  0.89  &  0.67  &  0.79  &  0.19  &  0.52  &  0.17  &  0.52  &  0.72  &  0.66  & 0.52 & 0.47$\pm$0.26 & 9.1 \\\hline 
\hline
 Median & 0.38  &  0.32  &  0.25  &  0.05  &  0.69  &  0.68  &  0.77  &  0.20  &  0.51  &  0.21  &  0.50  &  0.76  &  0.66  &  &  & \\\hline 
Average & 0.40  &  0.31  &  0.27  &  0.05  &  0.67  &  0.63  &  0.70  &  0.21  &  0.50  &  0.22  &  0.46  &  0.73  &  0.64  &  &  & \\\hline 
STD & 0.05  &  0.05  &  0.07  &  0.01  &  0.25  &  0.12  &  0.15  &  0.03  &  0.06  &  0.05  &  0.08  &  0.17  &  0.15  &  &  & \\\hline

    \end{tabular}
  \label{tblf}
\end{table*}

\begin{table*}[!t]
\setlength\tabcolsep{0.11cm}
\fontsize{9}{9}\selectfont
  \centering
  \caption{Median (sorted) Recall performance per dataset per benchmark}
    \begin{tabular}{>{\bfseries}llllllllllllll|ll}
    \hline
Method / Dataset & ANT & CML & IVY & jED & L4J & LUC & POI & PR6 & SYN & TOM & VEL & XAL & XER & Med. & Avg$\pm$Std \\\hline\hline

NNF-LOG &  0.31  &  0.20  &  0.35  &  0.64  &  0.16  &  0.36  &  0.18  &  0.26  &  0.26  &  0.39  &  0.23  &  0.29  &  0.19  & 0.26 & 0.29$\pm$0.12 \\\hline 
NNF-TunedLOG &  0.29  &  0.20  &  0.33  &  0.64  &  0.15  &  0.33  &  0.18  &  0.27  &  0.37  &  0.44  &  0.23  &  0.29  &  0.18  & 0.29 & 0.30$\pm$0.13 \\\hline 
NNF-NB &  0.45  &  0.14  &  0.45  &  0.45  &  0.07  &  0.21  &  0.17  &  0.48  &  0.45  &  0.75  &  0.23  &  0.30  &  0.16  & 0.30 & 0.33$\pm$0.19 \\\hline 
NNF-DT &  0.49  &  0.30  &  0.45  &  0.36  &  0.15  &  0.26  &  0.71  &  0.36  &  0.55  &  0.74  &  0.22  &  0.28  &  0.36  & 0.36 & 0.40$\pm$0.17 \\\hline 
NNF-TunedJ48 &  0.51  &  0.36  &  0.53  &  0.36  &  0.35  &  0.44  &  0.38  &  0.58  &  0.53  &  0.58  &  0.32  &  0.36  &  0.41  & 0.41 & 0.44$\pm$0.09 \\\hline 
NNF-J48 &  0.52  &  0.34  &  0.55  &  0.45  &  0.28  &  0.42  &  0.62  &  0.61  &  0.49  &  0.58  &  0.29  &  0.36  &  0.40  & 0.45 & 0.46$\pm$0.11 \\\hline 
NNF-TunedDT &  0.66  &  0.45  &  0.47  &  0.55  &  0.23  &  0.28  &  0.62  &  0.59  &  0.55  &  0.48  &  0.23  &  0.40  &  0.37  & 0.47 & 0.45$\pm$0.14 \\\hline 
NNF-TunedBN &  0.62  &  0.27  &  0.50  &  0.64  &  0.34  &  0.38  &  0.52  &  0.62  &  0.52  &  0.48  &  0.35  &  0.41  &  0.43  & 0.48 & 0.47$\pm$0.11 \\\hline 
LSH-J48 &  0.56  &  0.40  &  0.55  &  0.55  &  0.36  &  0.54  &  0.66  &  0.58  &  0.41  &  0.71  &  0.42  &  0.34  &  0.37  & 0.50 & 0.58$\pm$0.25 \\\hline 
LSH-TunedJ48 &  0.59  &  0.43  &  0.55  &  0.73  &  0.40  &  0.48  &  0.66  &  0.53  &  0.45  &  0.89  &  0.44  &  0.38  &  0.42  & 0.51 & 0.61$\pm$0.26 \\\hline 
GIS(VR-VMUL)-J48 &  0.61  &  0.40  &  0.56  &  0.50  &  0.35  &  0.92  &  0.54  &  0.55  &  0.51  &  0.84  &  0.49  &  0.48  &  0.44  & 0.52 & 0.55$\pm$0.18 \\\hline 
GIS(VR-VNN)-J48 &  0.61  &  0.48  &  0.61  &  0.64  &  0.37  &  0.49  &  0.66  &  0.59  &  0.50  &  0.68  &  0.46  &  0.44  &  0.44  & 0.52 & 0.54$\pm$0.14 \\\hline 
NNF-TunedNB &  0.88  &  0.40  &  0.70  &  0.64  &  0.22  &  0.55  &  0.40  &  0.73  &  0.67  &  0.83  &  0.47  &  0.38  &  0.30  & 0.55 & 0.55$\pm$0.20 \\\hline 
NNF-BN &  0.88  &  0.41  &  0.68  &  0.64  &  0.28  &  0.57  &  0.50  &  0.65  &  0.52  &  0.77  &  0.60  &  0.45  &  0.42  & 0.57 & 0.57$\pm$0.15 \\\hline 
GIS(VR-VMUL)-LOG &  0.75  &  0.49  &  0.60  &  0.64  &  0.56  &  0.87  &  0.77  &  0.34  &  0.63  &  0.62  &  0.48  &  0.66  &  0.58  & 0.62 & 0.61$\pm$0.16 \\\hline 
GIS(VR-VNN)-LOG &  0.77  &  0.52  &  0.60  &  0.64  &  0.43  &  0.66  &  0.81  &  0.73  &  0.62  &  0.73  &  0.56  &  0.58  &  0.51  & 0.64 & 0.63$\pm$0.13 \\\hline 
GIS(FX-VMUL)-LOG &  0.82  &  0.54  &  0.59  &  0.64  &  0.66  &  0.88  &  0.80  &  0.55  &  0.69  &  0.75  &  0.56  &  0.71  &  0.66  & 0.66 & 0.68$\pm$0.13 \\\hline 
GIS(FX-VNN)-LOG &  0.84  &  0.60  &  0.65  &  0.64  &  0.61  &  0.79  &  0.86  &  0.73  &  0.78  &  0.77  &  0.69  &  0.69  &  0.57  & 0.70 & 0.71$\pm$0.13 \\\hline 
GIS(FX-VMUL)-J48 &  0.88  &  0.59  &  0.74  &  0.64  &  0.56  &  1.00  &  0.80  &  0.70  &  0.92  &  0.94  &  0.51  &  0.67  &  0.64  & 0.72 & 0.73$\pm$0.17 \\\hline 
GIS(FX-VNN)-J48 &  0.94  &  0.74  &  0.79  &  0.64  &  0.49  &  0.71  &  0.85  &  0.92  &  0.70  &  0.85  &  0.70  &  0.67  &  0.55  & 0.75 & 0.74$\pm$0.16 \\\hline 
GIS(VR-VMUL)-DT &  0.81  &  0.96  &  0.78  &  1.00  &  1.00  &  0.59  &  0.73  &  0.70  &  0.69  &  0.81  &  0.94  &  0.55  &  0.42  & 0.77 & 0.74$\pm$0.19 \\\hline 
GIS(VR-VMUL)-BN &  0.95  &  0.91  &  0.78  &  0.64  &  1.00  &  0.44  &  0.95  &  0.61  &  0.99  &  0.77  &  0.96  &  0.55  &  0.37  & 0.78 & 0.76$\pm$0.22 \\\hline 
GIS(FX-VNN)-NB &  0.92  &  0.69  &  0.85  &  0.64  &  0.56  &  0.80  &  0.85  &  0.93  &  0.82  &  0.87  &  0.38  &  0.65  &  0.54  & 0.78 & 0.73$\pm$0.17 \\\hline 
GIS(VR-VMUL)-NB &  0.88  &  0.34  &  0.78  &  0.64  &  0.29  &  0.93  &  0.95  &  0.94  &  0.99  &  0.79  &  0.42  &  0.76  &  0.75  & 0.79 & 0.73$\pm$0.23 \\\hline 
GIS(FX-VMUL)-NB &  0.85  &  0.39  &  0.82  &  0.64  &  0.28  &  0.94  &  0.95  &  0.87  &  0.97  &  0.79  &  0.42  &  0.85  &  0.70  & 0.80 & 0.72$\pm$0.22 \\\hline 
GIS(VR-VNN)-NB &  0.95  &  0.79  &  0.85  &  0.64  &  0.67  &  0.83  &  0.88  &  0.92  &  0.85  &  0.91  &  0.79  &  0.65  &  0.57  & 0.83 & 0.79$\pm$0.15 \\\hline 
GIS(FX-VMUL)-DT &  0.96  &  0.96  &  0.75  &  1.00  &  0.88  &  0.89  &  0.79  &  0.84  &  0.78  &  0.81  &  0.94  &  0.64  &  0.51  & 0.86 & 0.81$\pm$0.18 \\\hline 
GIS(FX-VMUL)-BN &  0.96  &  0.96  &  0.79  &  0.64  &  1.00  &  0.47  &  0.95  &  0.91  &  1.00  &  0.80  &  0.96  &  0.63  &  0.48  & 0.88 & 0.80$\pm$0.19 \\\hline 
GIS(VR-VNN)-DT &  0.96  &  0.96  &  0.78  &  1.00  &  0.38  &  0.73  &  0.86  &  0.91  &  0.95  &  1.00  &  0.93  &  0.67  &  0.53  & 0.89 & 0.80$\pm$0.21 \\\hline 
GIS(FX-VNN)-DT &  0.96  &  0.96  &  0.75  &  1.00  &  0.85  &  0.89  &  0.94  &  0.91  &  0.98  &  1.00  &  0.94  &  0.72  &  0.57  & 0.91 & 0.84$\pm$0.20 \\\hline 
GIS(VR-VNN)-BN &  0.96  &  0.96  &  0.82  &  1.00  &  0.85  &  0.68  &  0.95  &  0.91  &  0.98  &  1.00  &  0.88  &  0.56  &  0.50  & 0.92 & 0.84$\pm$0.17 \\\hline 
GIS(FX-VNN)-BN &  0.96  &  0.96  &  0.80  &  1.00  &  0.85  &  0.73  &  0.95  &  0.91  &  0.98  &  0.97  &  0.92  &  0.68  &  0.54  & 0.94 & 0.87$\pm$0.14 \\\hline 
LSH-NB &  1.00  &  0.96  &  0.97  &  1.00  &  0.89  &  0.97  &  0.98  &  0.98  &  0.99  &  1.00  &  0.94  &  0.88  &  0.81  & 0.97 & 0.92$\pm$0.10 \\\hline 
LSH-TunedLOG &  1.00  &  1.00  &  1.00  &  1.00  &  0.98  &  1.00  &  1.00  &  1.00  &  0.99  &  1.00  &  0.97  &  0.90  &  0.88  & 1.00 & 0.94$\pm$0.13 \\\hline 
LSH-BN &  1.00  &  0.96  &  0.75  &  1.00  &  1.00  &  0.89  &  1.00  &  0.91  &  1.00  &  1.00  &  1.00  &  0.90  &  0.80  & 1.00 & 0.94$\pm$0.10 \\\hline 
LSH-DT &  1.00  &  0.96  &  1.00  &  1.00  &  1.00  &  1.00  &  0.95  &  0.91  &  0.62  &  1.00  &  1.00  &  0.99  &  1.00  & 1.00 & 0.93$\pm$0.18 \\\hline 
LSH-LOG &  0.99  &  1.00  &  1.00  &  1.00  &  1.00  &  0.98  &  1.00  &  1.00  &  0.99  &  1.00  &  0.97  &  0.91  &  0.84  & 1.00 & 0.95$\pm$0.10 \\\hline 
LSH-TunedBN &  1.00  &  0.96  &  0.89  &  1.00  &  1.00  &  0.38  &  1.00  &  0.91  &  1.00  &  1.00  &  1.00  &  1.00  &  0.50  & 1.00 & 0.86$\pm$0.23 \\\hline 
LSH-TunedDT &  1.00  &  0.96  &  1.00  &  1.00  &  1.00  &  1.00  &  0.96  &  0.91  &  0.50  &  1.00  &  1.00  &  0.99  &  1.00  & 1.00 & 0.88$\pm$0.22 \\\hline 
LSH-TunedNB &  1.00  &  1.00  &  1.00  &  1.00  &  1.00  &  1.00  &  1.00  &  1.00  &  1.00  &  1.00  &  1.00  &  0.82  &  0.90  & 1.00 & 0.97$\pm$0.08 \\\hline 
\hline
 Median & 0.88  &  0.60  &  0.75  &  0.64  &  0.56  &  0.72  &  0.84  &  0.74  &  0.75  &  0.82  &  0.60  &  0.62  &  0.52  &  &  \\\hline 
Average & 0.80  &  0.64  &  0.71  &  0.72  &  0.59  &  0.68  &  0.75  &  0.73  &  0.73  &  0.79  &  0.64  &  0.60  &  0.55  &  &  \\\hline 
STD & 0.21  &  0.29  &  0.20  &  0.22  &  0.32  &  0.25  &  0.25  &  0.22  &  0.24  &  0.19  &  0.29  &  0.22  &  0.22  &  &  \\\hline

    \end{tabular}
  \label{tblr}
\end{table*}

\subsection{Experiments and Abbreviations} 
Having multiple set of options for each benchmark, we identify them through abbreviations. 
These abbreviations include three main methods (\textbf{GIS}: Genetic Instance Selection, \textbf{LSH}: Locality Sensitive Hashing, \textbf{NNF}: NN-Filter), validation dataset selection methods (\textbf{VMUL}: multiple random validation datasets, \textbf{VNN}: NN-Filter based validation dataset), type of training data for GIS (\textbf{VR}: variable size, \textbf{FX}: fixed size). The additional tuning of the learners are specified in benchmark names as well. Hyper-parameter optimization of the learners is specified by using \textbf{``Tuned''} keyword appended to the tuned learners, e.g \textbf{TunedNB} means NB with optimization on multiple validation datasets. 

Please note that none of the GIS variants use hyper-parameter tuning in their base classifiers. We omitted GIS base learner optimization for the sake of runtime and practical considerations as the search method involves a large number of dataset evaluations for fitness assignments, making the optimization integration impractical using the current options. While variants of NNF and LSH use hyper-parameter optimization, we also report NNF and LSH variants without optimization to address the absence of optimization for GIS to some extent.

Overall, the three groups of approaches make 40 different benchmarks. For NN-Filter, 5 tuned and 5 non tuned learners are combined with NN-Filter (with tuned k), making 10 benchmarks. The same goes for the LSH (with tuned parameters), representing another 10 benchmarks. Despite not having learning technique optimization, GIS makes up for the rest of the benchmarks, i.e., the remaining 20. These benchmarks are generated by five learners, two validation methods (VMUL, VNN) and two training dataset types(VR, FX).

\section{Results}

\label{sec:res}

\begin{figure*}[!t]
    \centering
 
   \fbox{
   \includegraphics[width=0.98\textwidth,height=6.5cm, trim={1.4cm 3cm 1.8cm 4cm},clip]{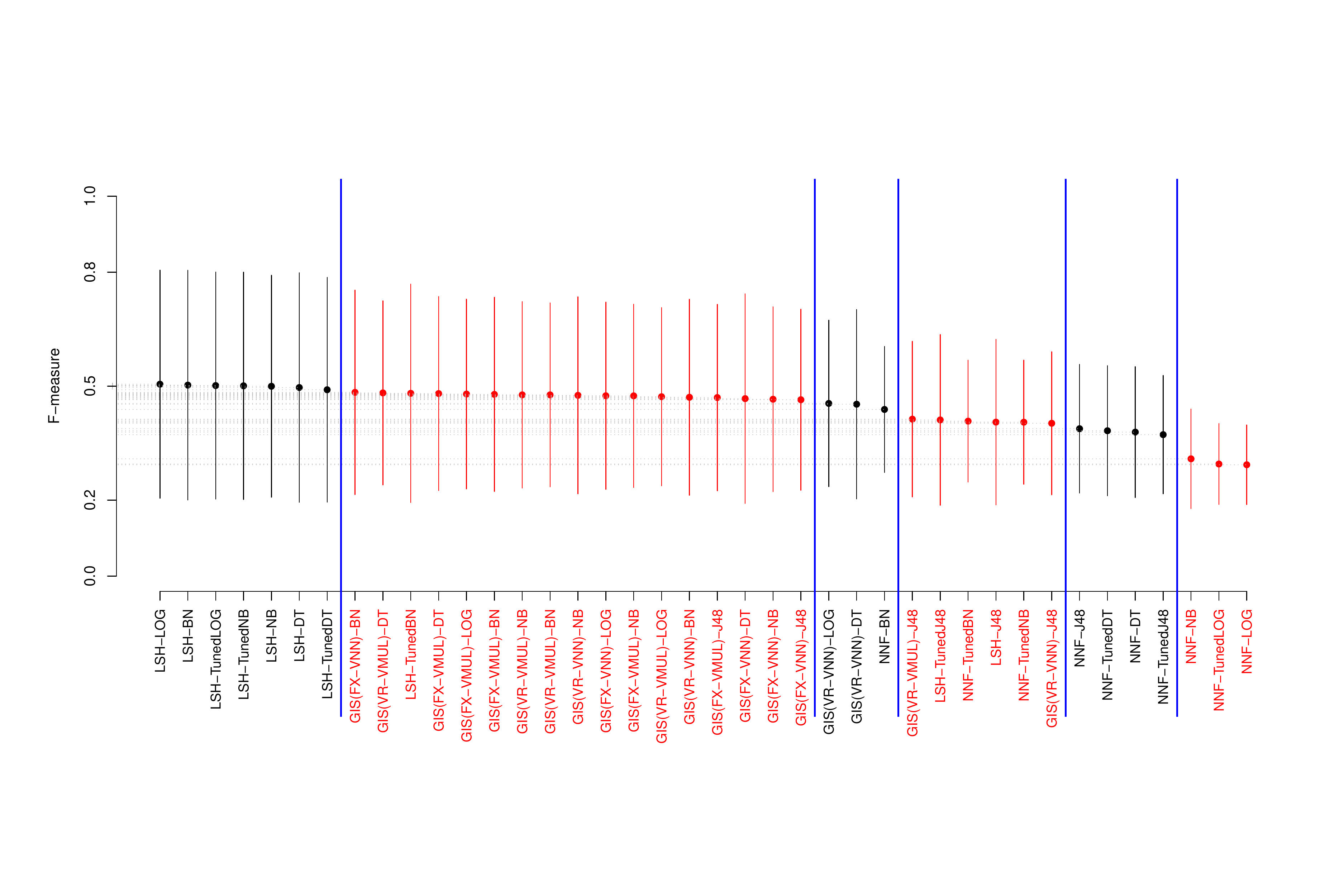}
   
   }
    
    \vspace{0.1cm}
    
    \fbox{
    \includegraphics[width=0.98\textwidth,height=6.5cm, trim={1.4cm 3cm 1.8cm 4cm},clip]{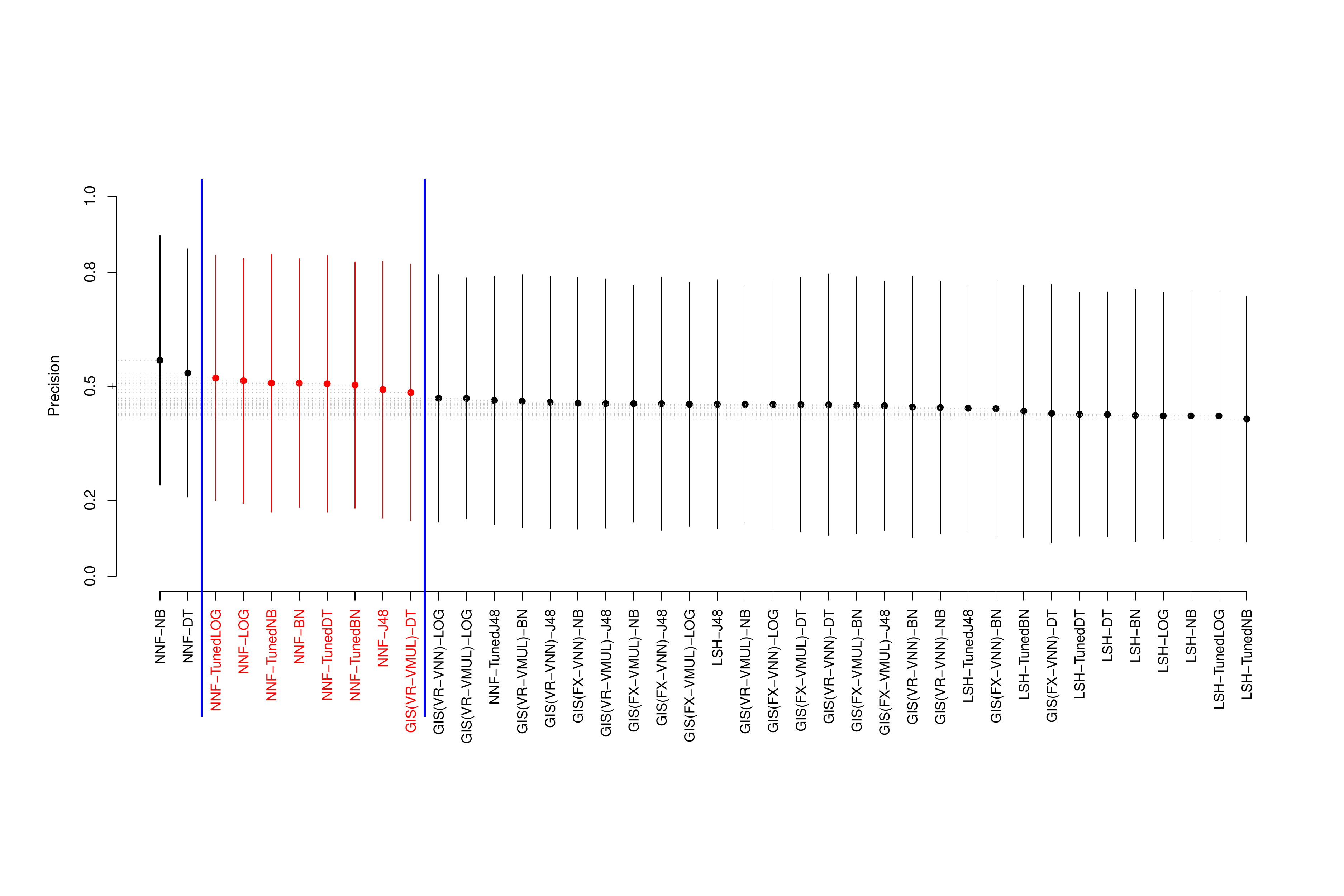}
   }
   
    \vspace{0.1cm}
    
    \fbox{
    \includegraphics[width=0.98\textwidth,height=6.5cm, trim={1.4cm 3cm 1.8cm 4cm},clip]{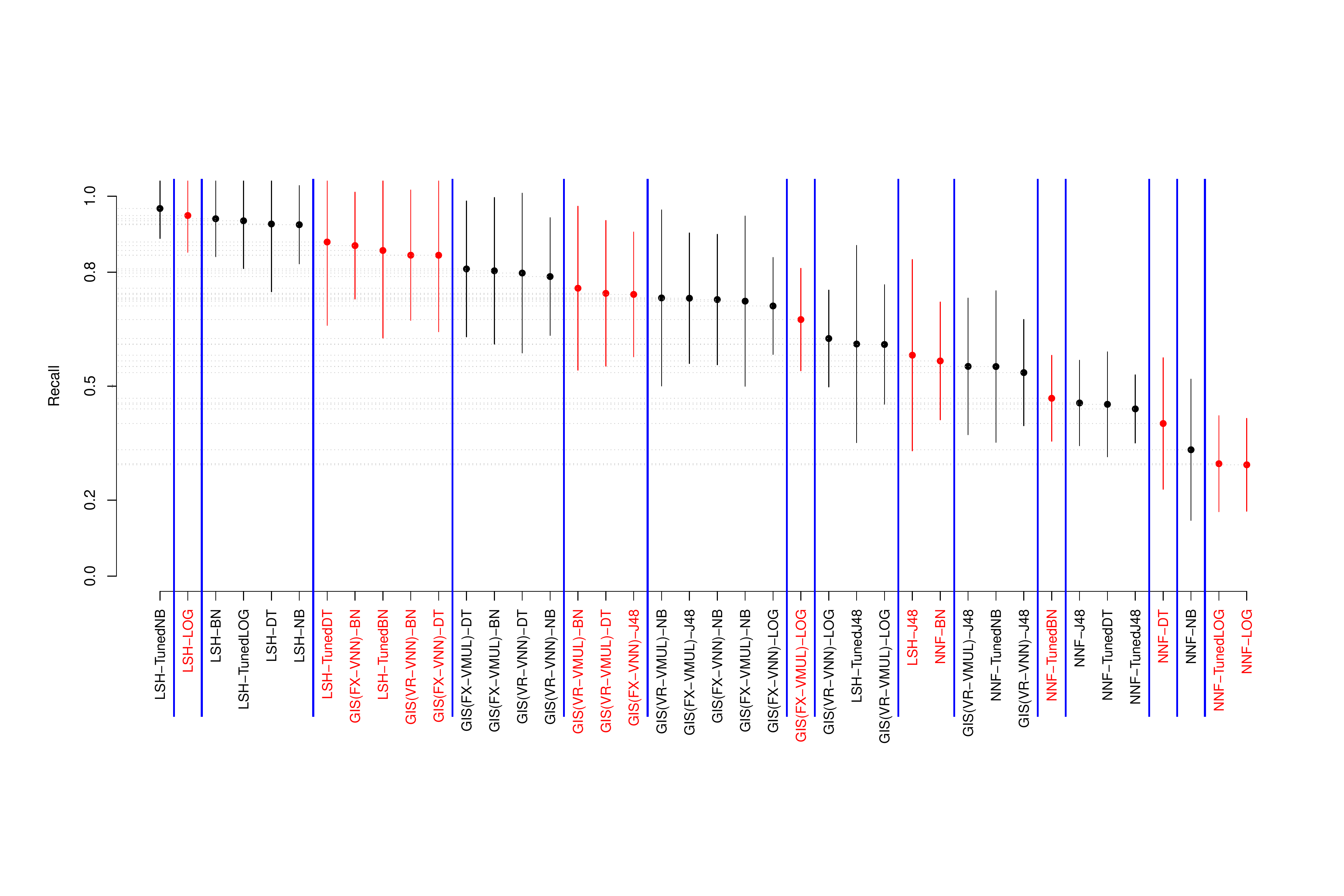}
    }
    
    \caption{The Scott-Knott Effect Size Difference (ESD) Tests for F-measure, Precision, Recall}
    \label{figsc}
\end{figure*}

Tables \ref{tblf} and \ref{tblr} provide the median values of the F-measure and recall, respectively, from the experiments performed for the CPDP benchmarks per benchmark and per dataset. Additional measures as well as the actual confusion matrices can be obtained from the replication package \cite{replpack}. The benchmarks are sorted increasingly top to bottom based on their overall median values in the tables, the top row having the lowest median value. Beside these tables, the results of the Scott-Knott ESD tests are presented through plots for F-measure, precision and recall in Figure \ref{figsc}. In Figure \ref{figsc}, distinct groups of performances are represented by two alternating colors (black and red). Neighboring benchmarks with different colors do not belong to the same group, i.e., statistically significant differences are observed. We separate the groups using long vertical blue color lines. The benchmarks within two blue lines belong to the same group. All the benchmarks between two long blue lines have the same color.

Median F-measure values in Table \ref{tblf} demonstrate  better performances of LSH and GIS benchmarks. While the majority of NNF variants show lower performances, NNF-BN, i.e., NN-Filter with Non Tuned Bayes Net as base learner, shows good performance. At the same time, TunedBN with NNF shows decreases in overall performance for the learner BN. In case of NNF, beside BN, for which negative effect of tuning is observed, slight decreases in performance can be seen for LOG, albeit insignificant. Overall, tuning does not seem to show a consistent behavior across the methods. For example, tuning shows positive effect on some of the LSH benchmarks (very small for LOG and J48), while the effect is significantly negative on some other, e.g., LSH-BN and LSH-TunedBN.

Different options for GIS tend to be related to higher overall median F-measure values. The top median F-measure performance belongs to GIS(VR-VNN)-BN, i.e., GIS with variable training dataset sizes and NN generated validation dataset and BN as base learner. Similar performances can be seen for GIS(FX-VNN)-BN (second highest median F-measure). Beside the aforementioned two and three GIS variants with DT learner, the majority of the other GIS variants show lower median F-measure values in comparison with different LSH variants.

For LSH, the highest median F-measure value is observed for LSH-TunedLOG and LSH-NB. LSH with Tuned Naive Bayes,  LSH with Logistic Regression and LSH with Bayes Net are next. While some good performances can be observed for tuned learners, one can not justifiably identify tuning as a factor for improving median F-measure performance, especially based on the ranking of the approaches, according to the statistical tests.

LSH with J48, i.e., LSH-J48 and LSH-TunedJ48, show the lowest median F-measure performances among LSH variants. LSH-TunedDT also show similar low performances while LSH-DT with better median F-measure is not among the best LSH counterparts as well. At the same time, BN and LOG and NB seem to be related to better LSH performances. In fact, NB appears often among the good performing approaches and this is not just limited to LSH. 

Except for one notable performance, i.e., GIS(FX-VNN)-J48, the rest of benchmarks with J48 as base learner are at the lower side of the performance spectrum. Notably, irrespective of the method, learner and tuning or lack thereof, the performances on JEdit-4.3 dataset are equally low for all benchmarks. This is due to the very high imbalance in the distribution of the defects in JEdit-4.3 dataset. The presence of only 2\% defective instances in JEdit-4.3 and the high difference with the rest of the datasets (training) in terms of defective instance rate leads to very low performances for JEdit. The low performances, however, are universal for all of the benchmarks.

The Scott-Knott ESD tests show six distinct groups for F-measure. The top group contains LSH benchmarks only while in the second best group all but one, i.e., LSH-TunedBN are GIS variants. Confirming the impact of learners, no variant of LSH with J48 are present at the top three groups while at least one benchmark appears at the top group for the other learners. The only benchmarks with J48 in the top three groups appear at the lower part of the second best group and are used in GIS as base learner. The only LSH methods in the second group uses learner tuning on BN, its non-tuned versions of which appear on the top group. 

NNF-BN, which is the only NNF variant in the top three groups, appears on the third group and is superior to the tuned version which is located in group four. The rest of the groups contain mostly other NNF benchmarks as well as LSH and GIS with J48. The last two groups contain only NNF variants. NNF-LOG, NNF-TunedLOG and NNF-NB are the only members of the last group, while their LSH counterparts all belong to the top ranking group. 

LSH and GIS favor recall more than precision and NNF benchmarks show better precision. In fact, only three significantly distinct groups of precision performances are detected by the tests, the first of which is comprised of NNF benchmarks only and the second group contains NNF variants except for GIS(VR-VMUL)-DT, the GIS variant with DT learner. Non of the other GIS and LSH variants seem to be significantly different from each other in terms of precision and they all are placed in the third group. The only NNF variant which are present at the lower performing group based on precision uses J48 as base learner. Recall wise, 16 different groups are identified. Of them, the top three groups contain only LSH methods, four of the next six contain GIS only and the last five contain only NNF. It is clear that the lower performances of LSH with J48 is recall based. BN appear more frequently than the rest of the learners at the top recall performing groups.

According to the Scott-Knott ESD tests (Figure \ref{figsc}), the NNF benchmarks never outperform the LSH counterparts with the same base learner.  In fact, LSH and NNF benchmarks are never parts of the same group if they use the same learner, meaning that LSH always achieves significantly better results than NNF with same settings. This is true even for J48, which tend to have the lowest performances among the used learners in terms of F-measure. The same is true for tuned learners in LSH and NNF as well as in for the majority of the cases for GIS. However, some of the LSH and GIS benchmarks belong to the same rank group, e.g., LSH-J48 and GIS(VR-VMUL)-J48, with GIS. 

In terms of the magnitude of the differences, the top group has very small differences with the second group ($d=0.0324$) and third group ($d=0.077$) and small to medium differences with the forth group ($d=0.149$) and medium to large difference with the last group ($d=0.354$). Similarly, second group outperforms the third group with very small differences ($d=0.057$) and has small to medium differences with the next two ($d=0.147$ and $d=0.198$) and medium to large difference with the last group ($d=0.404$). The last group shows significant decreases in performances in comparison with all groups according to the effect sizes.

The last column in Table \ref{tblf} presents the running time of each benchmarks in seconds for the times spent on generating the final instances for predictions. For NNF, it includes finding the optimal k value and for LSH it represents finding the optimal values for the parameters. As mentioned before, the runtimes for GIS necessarily include dataset evaluation for fitness assignment and generating new populations, which explains the generally higher runtimes. In addition, GIS with VNN contains an extra step of generating NN based validation datasets.  In any case, the LSH methods require less time than the search based approach and lower runtime than NN based method. This is the expected behaviour considering the approximate nature of the hash based method.

\section{Threats to Validity}
\label{sec:thr}
During an empirical study, one should be aware of the potential threats to the validity of the obtained results and derived conclusions \cite{A54}. The potential threats to the validity identified for this study are assessed in three categories, namely: construct, external and conclusion validity.

{\bf Construct validity:} The experimental datasets used in this study belong to a set of datasets collected from Java projects and can be subject to quality issues. This and our previous papers have proposed steps to address these issues. To address different aspects of the models, different strategies were used. Validation dataset selection methods, near neighbor selection limit method (k for NN-Filter), training dataset types, multiple learning techniques and learning technique hyper-parameter optimization methods are some of the steps taken to shed light on specific properties the approaches. 

{\bf External validity:} It is difficult to draw general conclusions from empirical studies of software engineering and our results are limited to the analyzed data and context \cite{A57}. Even though many researchers have used the same datasets as the basis of their conclusions, there is no assurance about the generalization of conclusions drawn from these projects. Particularly the applicability of the conclusions for commercial, proprietary and closed source software might be different.

{\bf Conclusion validity:} Our experiments are repeated 20 times to address the randomness and the results are compared using the Scott-Knott ESD  tests and relevant effect sizes. Another threat is the choice of the evaluation measures. Other researchers might consider different measures to evaluate the methods and as a consequence, some of the observations and conclusions may change. This issue is addressed through providing a replication package from which a diverse set of measures can be calculated.

\section{Discussion and Conclusion}
LSH is a suitable near neighbor selection approach for data mining operations on massive datasets. The nature of LSH, which is based on hashing, makes LSH a strong approach especially in situations where data are abundant. While within project data are usually rare in defect prediction, there are plenty of datasets collected from diverse set of projects which in turn can be used for defect prediction and are utilized by CPDP. Even though availability of data is very important, irrelevancy threatens the usefulness of data; an issue that is present in CPDP which is an instance of the dataset shift problem. This is why some studies in CPDP, including NN-Filter and GIS, have proposed filtering approaches to remove irrelevant data. 

Two problems need to be solved at this stage. First, the performance of NN-Filter is not very high. Second, GIS, which is built on top of NN-Filter, can be slow due to its search based nature and its inclusion of NN-Filter in one variant. This means that with large data NN-Filter becomes increasingly slow and GIS with NN-Filter will suffer as well. Additionally, tuning the learning technique hyper-parameters become increasingly difficult due to practical considerations for GIS and NN-Filter for larger datasets. These problems can be addressed by LSH based selections. 

One way to tackle this issue proposed in this study was to use a modified version of the grid search method. This approach solves one of the problems with GIS, i.e., the inclusion of a slow, exact neighbor search using NN-Filter as validation dataset selection method. The runtime requirements for this approach make the learning technique hyper-parameter optimization increasingly difficult. At the same time, increasing the population size will have very negative effect on runtime, considering a high increase in the number of required prediction for fitness assignments during the evolution process. We have kept these parameters small for feasibility reasons due to runtime overheads.

The approach to select the top performing bucket from the divisions created by LSH is based on the heuristic that a training dataset (among a set of candidate training datasets, e.g., the buckets) that has a higher performance on a set of other datasets (the multiple validation datasets in this paper) may potentially be more suitable to predict higher number of other unknown (test) datasets. On the other hand, potential irrelevant instances of data that might cause decreases in performance in other groups (buckets) are potentially mapped to similar clusters by LSH and, therefore, are not included in the process of predicting unknown (test) datasets. 

Based on the results achieved, the research question {\bf ``How is the performance of LSH near neighbor selection method, in comparison with the benchmark methods?"} is answered as follows: LSH variants significantly improve NN-Filter in terms of both effectiveness and runtime. They also outperform GIS in terms of runtime while achieving comparable or even better effectiveness performances. LSH favor recall more than precision. The highest degree of variability can be seen in terms of recall while the least is observed for precision. Finally, NN-Filter benchmarks never outperform the LSH counterparts with the same base learner, tuned and non-tuned, and they never even belong to the same rank group.

\bibliographystyle{unsrt}
\bibliography{sigproc.bib}

\end{document}